\documentclass{article} 
\usepackage{iclr2025_conference,times}

\usepackage{amsmath,amsfonts,bm}

\def\eqref#1{equation~\ref{#1}}

\def\1{\bm{1}}

\DeclareMathAlphabet{\mathsfit}{\encodingdefault}{\sfdefault}{m}{sl}
\SetMathAlphabet{\mathsfit}{bold}{\encodingdefault}{\sfdefault}{bx}{n}

\usepackage{hyperref}
\usepackage{url}
\usepackage[utf8]{inputenc} %
\usepackage[T1]{fontenc}    %
\usepackage{hyperref}       %
\usepackage{url}            %
\usepackage{booktabs}       %
\usepackage{amsfonts}       %
\usepackage{nicefrac}       %
\usepackage{microtype}      %

\usepackage{stmaryrd}
\usepackage{bbm}
\usepackage{threeparttable}
\usepackage{amsmath}
\usepackage[capitalize]{cleveref}
\usepackage{graphicx}
\usepackage{multirow}
\usepackage{xspace}
\usepackage{enumitem}
\usepackage{wrapfig}
\usepackage[ruled,linesnumbered]{algorithm2e}

\title{DEAL-YOLO: Drone-based Efficient Animal Localization using YOLO}

\author{Aditya Prashant Naidu \\
Dept. of Computer Science and Engineering\\
Manipal Institute of Technology\\
Manipal Academy of Higher Education\\
Manipal, Karnataka, India\\
\texttt{adityanaidu2004@gmail.com} \\
\And
Hem Gosalia\\
Dept. of Mechatronics Engineering\\
Manipal Institute of Technology\\
Manipal Academy of Higher Education\\
Manipal, Karnataka, India\\
\texttt{hem.gosalia3@gmail.com}
\And
Ishaan Gakhar\\
Dept. of Information and Communication Technology\\
Manipal Institute of Technology\\
Manipal Academy of Higher Education\\
Manipal, Karnataka, India\\
\texttt{ishaangakhar04@gmail.com}
\And
Shaurya Singh Rathore\\
Dept. of Data Science and Computer Applications\\
Manipal Institute of Technology\\
Manipal Academy of Higher Education\\
Manipal, Karnataka, India\\
\texttt{shauryarathore121@gmail.com}
\And
Krish Didwania\\
Dept. of Computer Science and Engineering\\
Manipal Institute of Technology\\
Manipal Academy of Higher Education\\
Manipal, Karnataka, India\\
\texttt{krishdidwania0674@gmailcom}
\And
Ujjwal Verma \\
Dept. of Electronics and Communication Engineering\\
Manipal Institute of Technology\\
Manipal Academy of Higher Education \\
Manipal, Karnataka, India\\
\texttt{ujjwal.verma@manipal.edu}
}

\iclrfinalcopy 
\begin{document}

\maketitle

\begin{abstract}
Although advances in deep learning and aerial surveillance technology are improving wildlife conservation efforts, complex and erratic environmental conditions still pose a problem, requiring innovative solutions for cost-effective small animal detection. This work introduces DEAL-YOLO, a novel approach that improves small object detection in Unmanned Aerial Vehicle (UAV) images by using multi-objective loss functions like Wise IoU (WIoU) and Normalized Wasserstein Distance (NWD), which prioritize pixels near the centre of the bounding box, ensuring smoother localization and reducing abrupt deviations. Additionally, the model is optimized through efficient feature extraction with Linear Deformable (LD) convolutions, enhancing accuracy while maintaining computational efficiency. The Scaled Sequence Feature Fusion (SSFF) module enhances object detection by effectively capturing inter-scale relationships, improving feature representation, and boosting metrics through optimized multiscale fusion.
Comparison with baseline models reveals high efficacy with up to 69.5\% fewer parameters compared to vanilla Yolov8-N, highlighting the robustness of the proposed modifications. Through this approach, our paper aims to facilitate the detection of endangered species, animal population analysis, habitat monitoring, biodiversity research, and various other applications that enrich wildlife conservation efforts. DEAL-YOLO employs a two-stage inference paradigm for object detection, refining selected regions to improve localization and confidence. This approach enhances performance, especially for small instances with low objectness scores. 

\end{abstract}

\section{Introduction and Previous Work}

Wildlife object detection has proven to be essential for all aspects related to biodiversity conservation. \citep{CHALMERS2021112,https://doi.org/10.1002/rse2.234,PENG2020364}. Accurate identification and tracking of animal species from aerial imagery allows the evaluation of population trends, habitat changes, and effective protection strategies. Traditional monitoring techniques such as ground surveys and camera trapping, can be hindered by their high costs and potential human biases \citep{https://doi.org/10.1111/brv.13152}. To this end, UAV's present a more efficient alternative, providing cost-effective, high-resolution aerial data with minimal human involvement. Recent advancements in deep learning have significantly enhanced the automation and quality of wildlife detection through Convolutional Neural Networks (CNNs) and object detection models \citep{AXFORD2024102842}. However, further improvements are needed to enhance performance on object detection while ensuring computational efficiency, particularly for deployment on UAV's.

Modern object detection models, particularly the You Only Look Once (YOLO) family \citep{Redmon2016,Bochkovskiy2020} and Faster R-CNN \citep{Ren2016}, have demonstrated superior accuracy in detecting and classifying objects in complex environments. However, wildlife detection presents unique challenges, particularly in UAV-based imagery. Small animal targets often occupy only a few pixels, making distinguishing them from the background difficult. In addition, occlusions, overlapped species, variations in lighting conditions, and environmental interference further complicate the detection process \citep{Eikelboom2019}. Recent advances in small object detection have introduced various techniques to improve accuracy, yet challenges persist in drone-based wildlife detection. RRNet \citep{Chen_2019_ICCV} employed AdaResampling for realistic augmentation but struggled with segmentation challenges in natural environments. RFLA \citep{10.1007/978-3-031-20077-9_31} assigned labels via Gaussian receptive fields but faced limitations with irregularly shaped animals. The Focus \& Detect framework \citep{KOYUN2022116675} enhanced small-object detection through high-resolution cropping but required extensive manual annotations. Cross-layer attention mechanisms \citep{9302755} amplified small object features but increased computational costs, while SSPNet \citep{9515145} fused multiscale features but diluted fine details. Wildlife detection models, such as those based on YOLOv5, YOLOv8, and Faster R-CNN \citep{rs16162929}, performed well on structured targets like livestock but struggled with camouflage and scale variations. CNN-based approaches for satellite imagery \citep{rs12122026} and Faster R-CNN with HRNet \citep{d14080624} improved small target recognition but suffered from anchor box limitations and false positives due to vegetation noise. Similarly, YOLOv6L \citep{CUSICK2024102707} detected static nests but was sensitive to resolution changes. Efficient object detection models have also been explored, with modifications to YOLOv5 \citep{app12147255} improving efficiency at the cost of fine-grained spatial details. UFPMP-Det \citep{Huang_Chen_Huang_2022} leveraged attention mechanisms but introduced computational overhead, while Drone-DETR \citep{s24175496} relied on large datasets and exhibited slow convergence. Efficient YOLOv7-Drone \citep{drones7100616} optimized UAV detection but struggled with camouflaged wildlife due to its reliance on accurate mask generation.

Despite these advancements, achieving robust and efficient wildlife detection for previous works is challenging for drone imagery due to limitations in feature resolution, fixed anchor boxes, and difficulty in distinguishing fine details amidst background noise. The main contributions of this work include:
\begin{itemize}
    \item \textbf{Optimization and Restructuring of YOLOv8}: Modifications introduced to  YOLOv8 such as efficient convolution modules and an optimized downsampling strategy, significantly reduce computational complexity while preserving high performance.
    \item \textbf{State-of-the-Art Performance at lower computational load}: Achieved superior detection accuracy with up to 69.6\% reduction in trainable parameters, effectively optimizing both efficiency and performance, thus showcasing applicability in real-world use cases.
    \item \textbf{Two-Stage Inference Strategy}: Introduced a novel and adaptive two-stage Region of Interest (RoI) based inference approach that enhances detection performance by refining bounding box predictions in ambiguous environments requiring fine-grained differentiation. This results in 4\% increase in Precision and 4.2\% increase in Recall on average.
\end{itemize}



\section{Proposed Methodology}
The proposed methodology utilizes a combination of advanced loss functions, architectural modifications, and inference strategies to enhance object detection performance in UAV imagery. In particular, DEAL-YOLO integrates the \textbf{Normalized Wasserstein Distance} \citep{wang2022normalizedgaussianwassersteindistance} to model bounding boxes as 2D Gaussian distributions, measuring the similarity between transformed predicted boxes and ground truth labels. By assigning greater importance to pixels near the center, this approach accounts for the smaller size of aerial objects and introduces smoothness to the bounding box deviations, with the Optimal Network theory underpinning the exponential normalization to yield an effective similarity measure. To further mitigate the influence of low-quality examples, the model also incorporates the \textbf{Wise IoU} metric \citep{tong2023wiseiouboundingboxregression}, which minimizes the adverse effects of geometric variations—such as differences in distance and aspect ratio—by penalizing both major and minor misalignments between predicted anchor boxes and target boxes. Its adaptive weighting mechanism is particularly valuable for UAV applications, where altitude variations cause objects to appear at diverse scales, ensuring that smaller objects (often captured at high altitudes) are detected with improved precision. This combined approach is the first of its kind to be leveraged in the domain of UAV-based detection for accurate prediction and ensuring robust performance in complex aerial environments. The mathematical formulation of the same is detailed in the appendix.

\begin{figure}
    \centering
    \includegraphics[width=\textwidth, height=2.5in]{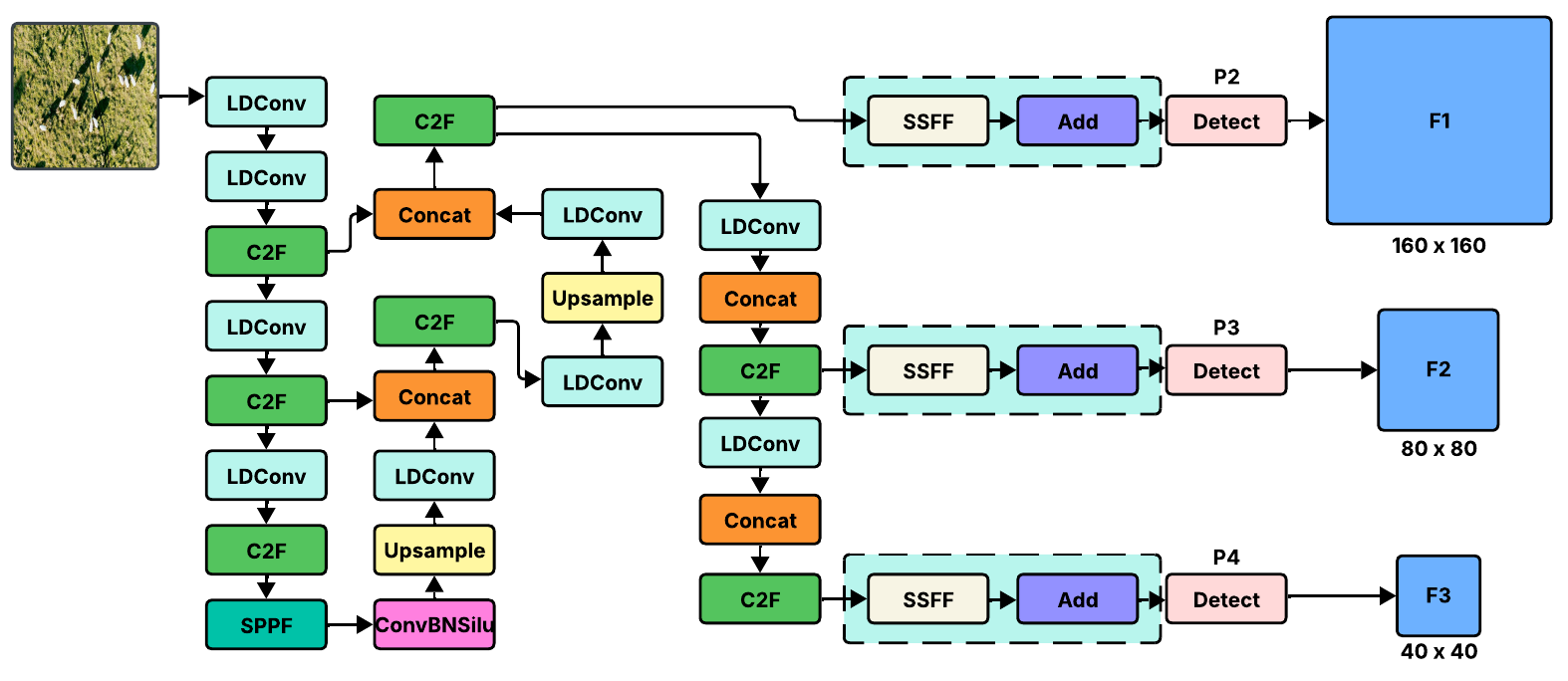}
    \caption{Schematic overview of the proposed model. Our contributions to the YOLOv8 model are highlighted in Cyan. F1, F2, and F3 represent the feature maps with their corresponding dimensions. All other blocks are taken directly from YOLOv8 \citet{yolov8_ultralytics}.}
    \label{fig:model_diagram}
\end{figure}

Within the YOLO framework, the \textbf{Feature Pyramid Network} (FPN) produces feature maps at multiple scales, typically designated as P2 and P5 \citep{wang2022yolov7trainablebagoffreebiessets}. While P2, a shallow layer with a smaller receptive field, captures fine, high-resolution details ideal for detecting small objects, P5, with its larger receptive field and coarser features, is more suited for large objects. As seen in Fig. \ref{fig:model_diagram}, the computational complexity is optimized  by excluding the P5 scale feature map from both the backbone and the FPN with a slight trade-off in performance. Consequently, the number of channels in the SPPF (Spatial Pyramid Pooling-Fast) blocks is reduced from 1024 to 512, enhancing feature extraction by focusing on the most relevant maps for the task of UAV detection.


Additionally, as seen in Fig. \ref{fig:model_diagram}, the \textbf{SSFF module} \citep{Kang_2024} is incorporated to enhance the extraction of multiscale information. Traditional fusion methods, such as simple summation or concatenation, often fall short of capturing complex inter-scale relationships. The SSFF module addresses this by normalizing, upsampling, and concatenating multiscale features into a 3D convolutional structure, which effectively handles objects with varying sizes, orientations, and aspect ratios. This multi-scale fusion is especially beneficial in UAV applications, where targets frequently exhibit diverse spatial characteristics and appear at different scales due to varying altitudes and camera angles. Moreover, the integration of \textbf{Linear Deformable (LD) convolutions} \citep{ZHANG2024105190} further refines feature extraction by dynamically adapting convolutional kernels based on local feature variations, thereby accommodating the geometric distortions and irregular shapes often observed in aerial imagery. This combination lightens the model, reduces computational overhead, and maintains competitive detection performance, making it particularly well-suited for UAV-based object detection tasks.

Finally, our methodology includes a \textbf{two-stage inference} approach, termed confidence-guided adaptive refinement, to improve detection accuracy, particularly for low-confidence detections. The first stage produces preliminary detections on the full-resolution image. Detections with a confidence score below a specified threshold are then refined in a second pass via adaptive region cropping, which extracts and resizes candidate regions relative to a high-confidence reference, resulting in an increased confidence score. The refined detection coordinates are transformed back to the scale of the original image, and Non-Maximum Suppression (NMS) is applied to remove duplicates. This dual-stage process balances computational efficiency and accuracy by concentrating refinement efforts on the most uncertain detections, thereby assimilating global context and local details to optimize performance. Overall, these combined strategies contribute to a robust and efficient detection pipeline tailored for UAV imagery, particularly in challenging environments such as wildlife detection.

\begin{table}[h]
  \caption{Comparisons with baseline YOLO models on the BuckTales dataset against the proposed approach across various metrics. Suffix 'T' stands for Tiny and 'N' stands for Nano. The '*' represents results with 2-stage inference.}
  \label{BuckTales_Patched_Results}
  \small
  \begin{center}
  \begin{tabular}{lcccccc}
    \toprule
    \textbf{Model} & \textbf{\#Params(M)} & \textbf{Precision} & \textbf{Recall} & \textbf{mAP$_{50}$} \\
    \midrule
    YOLOv5-N & 2.504  & 46.9  & 53.1  & 48.7  \\
    YOLOv6-N & 4.234  & 38.7  & 42.2  & 42.3  \\
    YOLOv8-N & 3.006  & 70.7  & 41.8  & 42.8  \\
    YOLOv9-T & 1.972  & 59.7  & 48.8  & 55.8  \\
    YOLOv10-N & 2.697  & 42.0  & 45.7  & 46.2  \\
    Gold-YOLO & 5.610  & 38.6  & 75.0  & 50.7  \\
    RT-DETR & 41.97  & 48.1  & 38.9  & 35.7  \\
    Faster-RCNN & 43.060  & 63.6  & 75.2  & 29.7  \\ 
    \hline
    DEAL-YOLO (Ours) & \textbf{0.994} & 75.3 & 58.2 & \textbf{48.5} \\
    DEAL-YOLO (Ours)* & \textbf{0.994} & \textbf{85.3} & \textbf{87.8} & 47.6 \\
    \bottomrule
  \end{tabular}
  \end{center}
\end{table}

\begin{table}[h]
\small
\caption{Comparison of SOTA methods against the proposed method on the WAID dataset across various metrics. Not all models have published the mAP$_{50}$, and hence that entry has been left blank. Suffix 'T' stands for Tiny, 'S' stands for Small and 'N' stands for Nano. The '*' represents results with 2-stage inference. '-LD' represents our model with LD convolutions.}
\label{WAID_Results}
\begin{center}
\begin{tabular}{lccccc}
\toprule
\textbf{Model} & \textbf{\#Params(M)} & \textbf{Precision} & \textbf{Recall} & \textbf{mAP$_{50}$} \\
\midrule
YOLOv7-T     & 6.000   & 93.7   & 92.3   & 95.23  \\
YOLOv5-S     & 7.200   & \textbf{96.9}   & 92.9   & \textbf{96.3}   \\
ADD-YOLO     & 1.500   & 93.0   & 91.0   & 95.0   \\
WILD-YOLO    & 12.380  & 92.8   & 91.41  & 95.0   \\
YOLOv4-S     & 8.270   & 38.2   & 92.7   & 56.3   \\
MobileNet v2 & 3.950   & 40.0   & 91.5   & 59.1   \\
YOLOv8-N     & 3.010   & 88.5   & 85.4   & 89.7   \\
\hline
DEAL-YOLO-LD   & \textbf{0.914}  & 91.1 & 88.9 & 93.4  \\
DEAL-YOLO      & 0.994  & 92.0 & 88.8 & 93.3  \\
DEAL-YOLO-LD*  & \textbf{0.914}  & 95.2 & 94.8 & 90.5  \\
DEAL-YOLO*     & 0.994  & 95.9 & \textbf{95.3} & 90.8  \\
\bottomrule
\end{tabular}
\end{center}
\end{table}

\section{Experiments and Results}


To validate the proposed methodology, the WAID \citep{app131810397} and BuckTales datasets \citep{naik2024bucktalesmultiuavdataset} have been employed and exhaustive experimentation has been performed to justify our choice of modules.

As evident in Table \ref{BuckTales_Patched_Results}, comparisons are drawn between various baseline YOLO models, specifically YOLOv6, YOLOv8, YOLOv9, YOLOv10, Gold-YOLO, RT-DETR, Faster-RCNN\citep{li2023yolov6, yolov8_ultralytics, wang2024yolov9, THU-MIGyolov10, wang2023goldyoloefficientobjectdetector, zhao2024detrsbeatyolosrealtime, Ren2016} and DEAL-YOLO. Here, we show the performance of our model at a much lower computational load. With a 68\% reduction in parameters, our model outperforms these baselines by an average of 4.8\% across all metrics. This drop in computational load, along with superior performance, makes our model suitable for the task of Animal Detection. As noted in the same table, we have presented results without the 2-stage inference, which are notably lower, further emphasizing the improvement brought in by our methodology. The proposed model was trained using the SOAP optimizer \citep{vyas2025soapimprovingstabilizingshampoo}, which offers better stability and convergence compared to Adam.

In Table \ref{WAID_Results}, the proposed methodology demonstrates performance comparable to SOTA, at \textbf{87\% less} parameters than YOLOv8n.Compared to YOLOv7-T, YOLOv5-S, ADD-YOLO, WILD-YOLO, YOLOv4-S, MobileNet v2, and YOLOv8-N, DEAL-YOLO LD, with its SSFF layer and LD convolutions, maintains strong performance in predicting bounding boxes, particularly for smaller objects such as the animals in the WAID dataset. Moreover, the advantages of 2-stage inference are clearly demonstrated, reflecting an improvement over standard inference for both DEAL-YOLO LD and DEAL-YOLO, even when accounting for the inclusion of lower confidence predictions. Additional experiments and details are mentioned in the appendix.




\section{Conclusion}

In this work, we have presented DEAL-YOLO, a novel approach to animal detection that showcases the superior performance of up to at 66.93\% lesser trainable parameters on BuckTales and comparable metrics to SOTA at 69.59\% lesser trainable parameters on WAID. 
The SSFF module, LD convolutions, and our novel 2-stage inference setup demonstrate excellent results across UAV-captured datasets like WAID and BuckTales.

\section{Acknowlegdement}

We would like to thank Mars Rover Manipal, an interdisciplinary student project of MAHE, for providing the essential resources and infrastructure that supported our research. We also extend our gratitude to Mohammed Sulaiman for his contributions in facilitating access to additional resources crucial to this work.

\bibliography{iclr2025_conference}
\bibliographystyle{iclr2025_conference}

\newpage
\appendix

\section{Detailed Methodology}

The Scaled Sequential Feature Fusion (SSFF) \citep{Kang_2024} block enhances multi-scale feature representation by refining feature maps (P3, P4, P5) sequentially using Gaussian smoothing before fusion. Each feature map \( f(i, j) \) is convolved with a Gaussian kernel \( G_{\sigma}(x, y) \), defined as:  

\begin{equation}
F_{\sigma}(i, j) = \sum_{u} \sum_{v} f(i - u, j - v) \times G_{\sigma}(u, v)
\end{equation}

where \( G_{\sigma}(x, y) \) is given by:  

\begin{equation}
G_{\sigma}(x, y) = \frac{1}{2\pi\sigma^2} e^{-\frac{x^2 + y^2}{2\sigma^2}}
\end{equation}

This progressively smooths the feature maps with increasing standard deviation \( \sigma \), ensuring robust feature refinement. The smoothed feature maps are then fused sequentially, allowing finer-scale details from P3 to progressively enhance coarser features in P4 and P5, preserving spatial relationships and improving object detection across scales.

We use the Normalized Wasserstein Distance to achieve smoothness of the bounding box deviations according to the formula
\begin{equation}
NWD(N_a, N_b) = \exp \left( -\frac{\sqrt{W_2^2(N_a, N_b)}}{C} \right),
\end{equation}
Where \(N_a\) and \(N_b\) represent two Gaussian distributions. The term \(W_2^2(N_a, N_b)\) denotes the squared 2-Wasserstein distance between these distributions, measuring the optimal transport cost between them. The constant \(C\) serves as a normalization constant.

This is combined with the Wise IoU metric to minimize the adverse effects of geometric variation according to
\begin{equation}
\mathcal{L}_{WIoU} = \mathcal{R}_{WIoU} \mathcal{L}_{IoU}
\end{equation}

\begin{equation}
\mathcal{R}_{WIoU} = \exp \left( \frac{(x - x_{gt})^2 + (y - y_{gt})^2}{(W_g^2 + H_g^2)^{\star}} \right)
\end{equation}
In the equations, \(\mathcal{L}_{WIoU}\) denotes the weighted IoU loss (computed as the product of the standard IoU loss \(\mathcal{L}_{IoU}\) and the scaling factor \(\mathcal{R}_{WIoU}\)). \(\mathcal{R}_{WIoU}\) scales the loss based on the squared Euclidean distance between the predicted box centre \((x,y)\) and the ground truth centre \((x_{gt},y_{gt})\), with \((W_g^2 + H_g^2)^{\star}\) serving as a normalization term.

LDConv (Linear Deformable Convolution) introduced by \citet{ZHANG2024105190} is an innovative convolutional operation that facilitates arbitrarily sampled shapes and accommodates a flexible number of parameters, distinguishing it from conventional fixed-grid convolutions. This approach generates initial sampled positions and learns offsets to adjust the receptive field dynamically. As a result, it enables more efficient feature extraction with linear parameter growth. This adaptability allows LDConv to cater to various target shapes while optimizing computational efficiency.

\begin{figure}[t]
    \centering
    \includegraphics[width=\textwidth,height=4in]{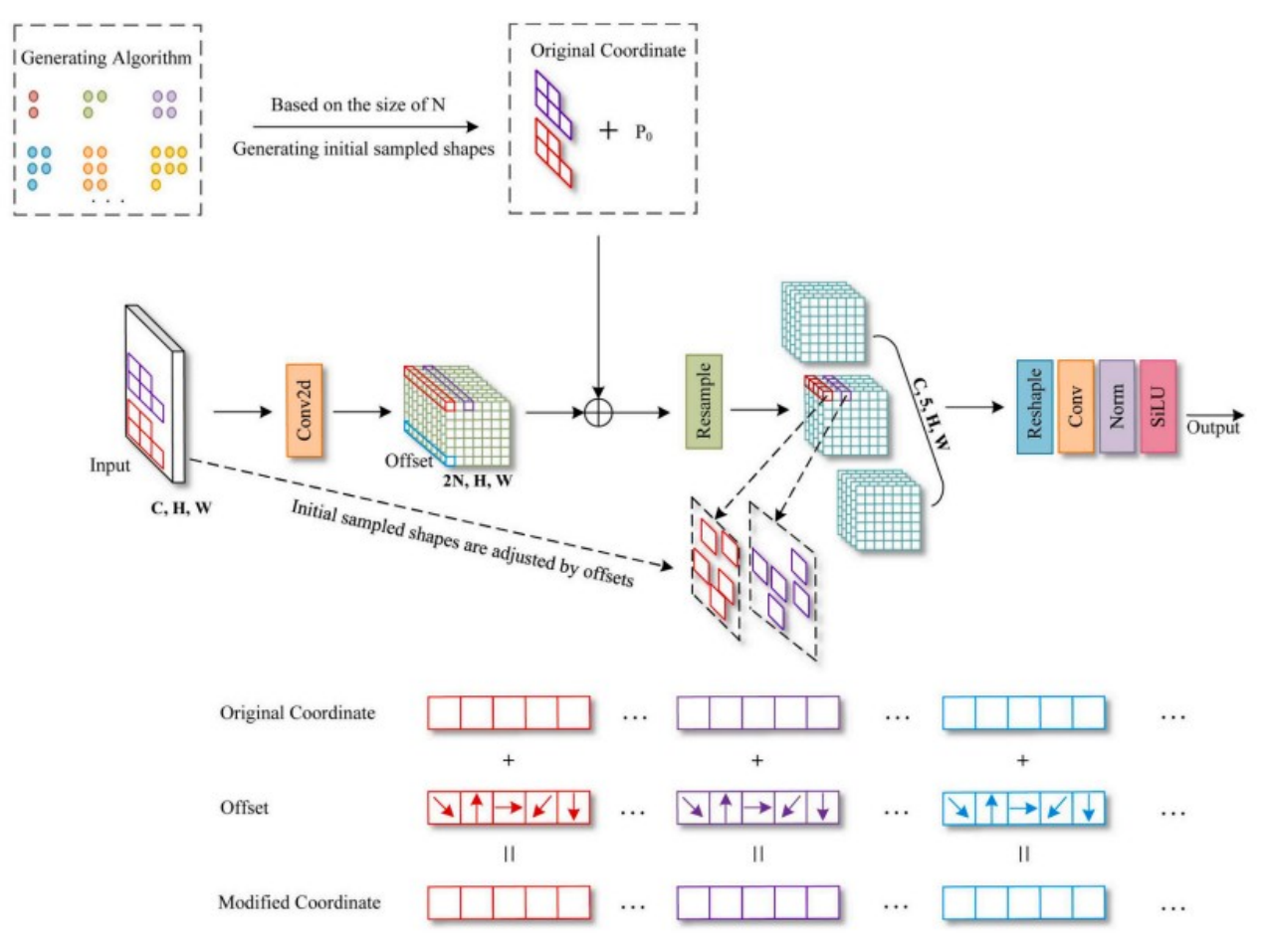}
    \caption{Schematic overview of the structure of LDConv.\citep{ZHANG2024105190} The initial sampled coordinates are assigned to a convolution of arbitrary size, and the sample shape is adjusted using learnable offsets. This process modifies the original sampled shape at each position through resampling.}
    \label{fig:LDConv_diagram}
\end{figure}

\section{Ablation Study}

In this study, the impact of individual components and their combination is presented for each dataset. As seen in Table \ref{Ablation_Study1}, the results of the SOAP optimizer are seen in Row 2. The results of Row 1 are Vanilla YOLOv8 \citep{yolov8_ultralytics} with the Adam Optimizer \citep{kingma2017adammethodstochasticoptimization}. Inclusion of the SOAP optimizer results in a 7\% increase while integrating the SSFF module along with WIoU and NWD Loss demonstrates an increase in performance of 6.625\%. Finally, changing P5 to P2 causes a 66.93\% reduction in trainable parameters while having a negligible effect on the quantitative metrics. These results showcase the effectiveness and fine-grained choice of our modules and the applicability of our approach in real-world scenarios. 

\begin{table}[h]
\caption{Ablation study on the impact of individual proposed changes in YOLOv8-N for the BuckTales Dataset. 'WIoU + NWD' represents the integration of WIoU and NWD losses.}
\label{Ablation_Study1}
\begin{center}
\setlength{\tabcolsep}{2.5pt}
\begin{tabular}{cccccccccc}
    \toprule
    & \textbf{SOAP}  & \textbf{SSFF}  & \textbf{WIoU+NWD}  & \textbf{P2 for P5} &  \textbf{\#Params(M)}  & \textbf{Precision}  & \textbf{Recall}  & \textbf{mAP$_{50}$}  & \textbf{AP$_{50-95}$}  \\ 
    \hline
    \specialrule{0em}{0.8pt}{0.8pt}
    \checkmark  &  &  &  &  & 3.006  & 92.2  & 87.7  & 93.1  & 61.0  \\
    \checkmark & \checkmark &  &  &  & 2.490  & 91.4  & 89.0  & 93.6  & \textbf{68.0}  \\
    \checkmark & \checkmark & \checkmark  &  &  & 2.490  & \textbf{94.4}  & \textbf{89.4}  & \textbf{94.2}  & 60.9  \\
    \checkmark & \checkmark & \checkmark  & \checkmark &  & 0.994  & 92.0  & 88.8  & 93.3  & 60.9  \\
    \checkmark & \checkmark & \checkmark  & \checkmark  & \checkmark & \textbf{0.914}  & 91.1  & 88.9  & 93.4  & 60.3  \\
    \bottomrule
\end{tabular}
\end{center}
\end{table}

\begin{table}[h]
\caption{Ablation study on the impact of individual proposed changes for WAID Dataset. 'WIoU + NWD' represents the integration of WIoU and NWD losses.}

\label{Ablation_Study2}
\begin{center}
\setlength{\tabcolsep}{2.5pt}
\begin{tabular}{cccccccccc}
    \toprule
    & \textbf{SOAP}  & \textbf{SSFF}  & \textbf{WIoU+NWD}  & \textbf{P2 for P5} &  \textbf{\#Params(M)}  & \textbf{Precision}  & \textbf{Recall}  & \textbf{mAP$_{50}$}  & \textbf{AP$_{50-95}$}  \\ 
    \hline
    \specialrule{0em}{0.8pt}{0.8pt}
    &  &  &  &  & 3.006  & 70.7  & 41.8  & 42.8  & 26.0  \\
    & \checkmark  &  &  &  & 3.006  & 62.4  & 57.1  & 56.6  & 33.2  \\
    & \checkmark  & \checkmark  &  &  & 2.490  & 65.4  & 53.7  & 58.0  & 33.0   \\
    & \checkmark  & \checkmark  & \checkmark  &  & 2.490  & \textbf{73.4}  & \textbf{60.7}  & \textbf{63.1}  & \textbf{39.4}  \\
    & \checkmark  & \checkmark  & \checkmark  & \checkmark  & \textbf{0.994}  & 67.4  & 57.5  & 63.0  & 37.6  \\
    \bottomrule
\end{tabular}
\end{center}
\end{table}



\renewcommand{\arraystretch}{1.3} 

\begin{table}[h]
  \caption{Ablation study on the effect of using the patched/unpatched versions of the BuckTales dataset, as well as different image resizing during inference. 'Patched/1280' means patched dataset and images were resized to 1280 during inference. Suffix 'T' stands for Tiny and 'N' stands for Nano.}
  \label{BuckTales_Resize_Exp}
  \small
  \begin{center}
  \begin{tabular}{lcccccc}  
    \toprule
    \textbf{BuckTales} & \textbf{Model} & \textbf{\#Params(M)} & \textbf{Precision} & \textbf{Recall} & \textbf{mAP$_{50}$} & \textbf{mAP$_{50-95}$} \\
    \midrule
    Patched/1280 
    & YOLOv5-N & 2.504  & 58.6 & 63.0 & \textbf{65.9} & 35.0   \\
    & YOLOv6-N & 4.234  & 54.9 & 61.6 & 61.6 & 34.6   \\
    & YOLOv8-N & 3.006  & 57.5 & 54.5 & 59.7 & 34.7   \\
    & YOLOv9-T & 1.972  & 57.3 & \textbf{66.0} & 65.1 & \textbf{40.1}   \\
    & YOLOv10-N & 2.697  & \textbf{63.7} & 55.1 & 57.6 & 33.5   \\
    & DEAL-YOLO (Ours) & \textbf{0.994} & 57.5 & 61.0 & 62.2 & 38.3 \\
    \midrule
    Unpatched/2560 
    & YOLOv5-N & 2.504  & 9.00 & 10.00 & 4.58 & 1.45   \\
    & YOLOv6-N & 4.234  & 7.87 & 8.93 & 3.68 & 1.15   \\
    & YOLOv8-N & 3.006  & 10.30 & 3.46 & \textbf{5.73} & \textbf{1.59}   \\
    & YOLOv9-T & 1.972  & 9.39 & 11.10 & 5.21 & 1.69   \\
    & YOLOv10-N & 2.697  & 5.41 & \textbf{14.20} & 3.36 & 1.16   \\
    & DEAL-YOLO (Ours) & \textbf{0.994} & \textbf{32.50} & 7.34 & 7.82 & 2.72 \\
    \midrule
    Unpatched/3840 
    & YOLOv5-N & 2.504  & 28.7 & 28.7 & 28.4 & 11.4   \\
    & YOLOv6-N & 4.234  & 47.8 & 29.0 & 28.9 & 11.6   \\
    & YOLOv8-N & 3.006  & \textbf{51.6} & 27.6 & 30.7 & 13.8   \\
    & YOLOv9-T & 1.972  & 48.0 & 35.3 & 36.8 & 15.4   \\
    & YOLOv10-N & 2.697  & 27.2 & 26.7 & 25.2 & 9.8   \\
    & DEAL-YOLO (Ours) & \textbf{0.994} & 40.0 & \textbf{39.1} & \textbf{38.5} & \textbf{19.0} \\
    \bottomrule
  \end{tabular}
  \end{center}
\end{table}

The decision to resize images to 640 at inference was based on practical considerations and the need for consistency with the patched dataset. Since the training was conducted on patched images (608$\times$513) resized to 1280, it was important to ensure that inference-time resizing did not introduce distortions that could impact model performance.

Resizing test images to 640 provides a reasonable balance between preserving spatial information and maintaining consistency with the training data. Since the patched images used during training are relatively small (608$\times$513), a test resolution of 640 minimizes excessive resizing, helping retain object details and prevent artifacts. As observed in Table \ref{BuckTales_Resize_Exp}, models evaluated on patched images (resized to 1280) achieve significantly better performance compared to those tested on unpatched images with larger resizing scales.

Additionally, using 640 aligns closely with standard YOLO input sizes, ensuring compatibility with pre-trained backbone architectures while keeping computational requirements manageable. Since UAV imagery contains small objects across large backgrounds, aggressive resizing (either upscaling or downscaling) could lead to a loss of fine details or unnecessary blurring.

While the exact impact of different test-time resizing strategies would require further empirical validation, the choice of 640 appears to be a well-reasoned approach that maintains consistency with training, minimizes distortions, and balances computational efficiency without introducing significant domain shifts.

\section{Qualitative Results}

\begin{figure}[h]
    \centering
    \includegraphics[width=5in,height=3in ]{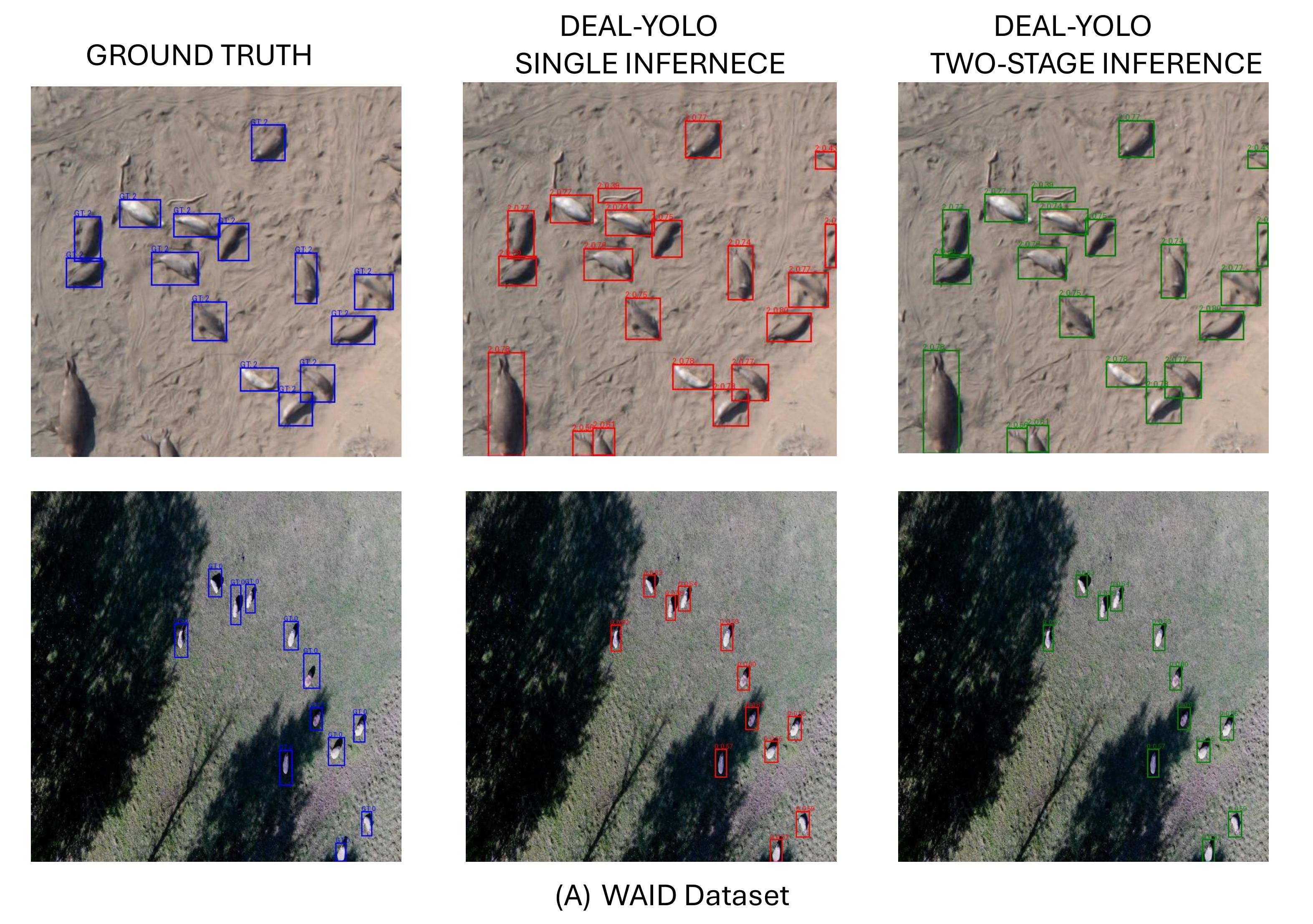} \\
    \includegraphics[width=5in,height=3in]{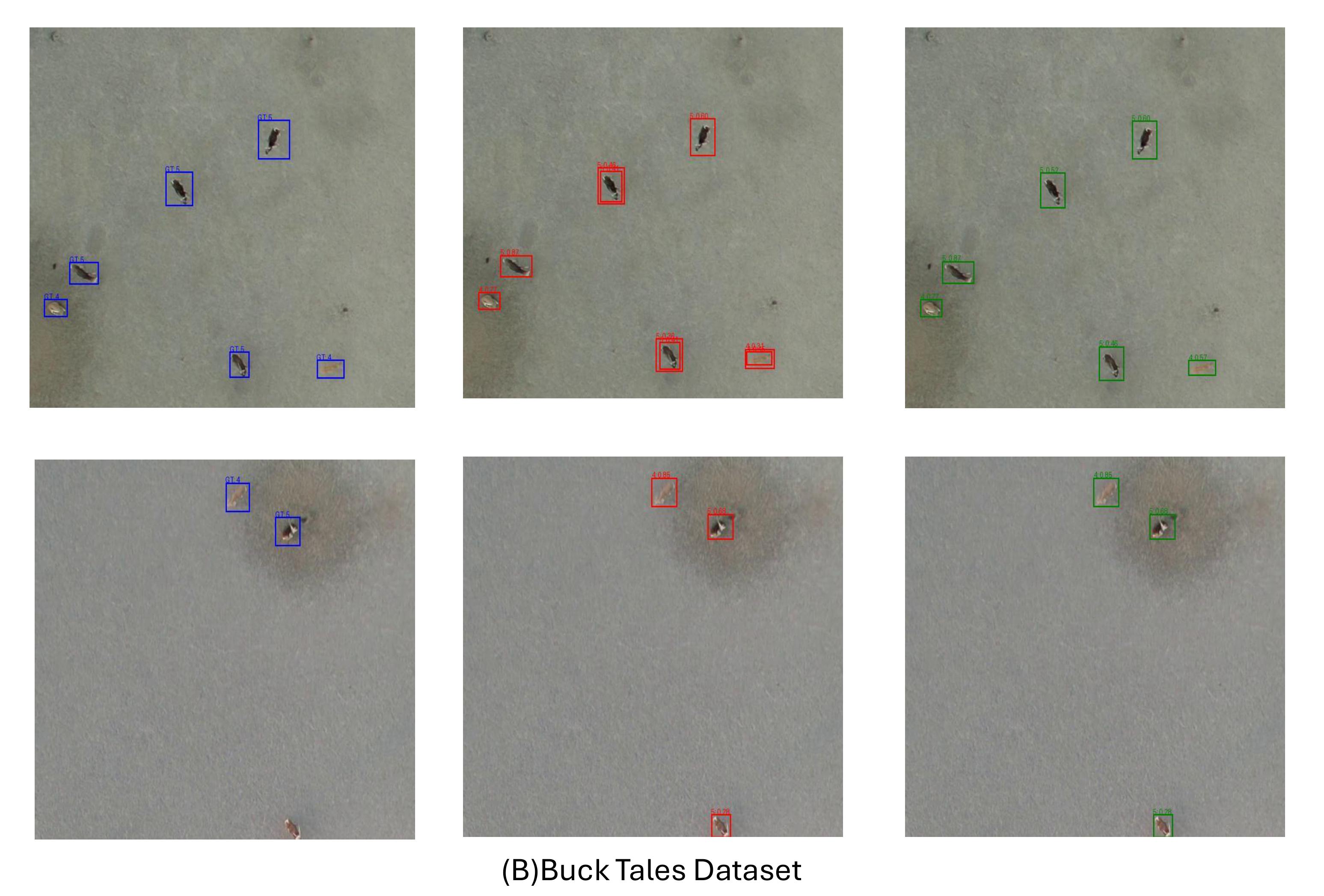} 
    \label{fig:combined}
    \caption{Qualitative results on the WAID and BuckTales datasets. Ground truth annotations are shown in blue, single-stage inference predictions in red, and two-stage inference predictions in green. The left column represents the Ground Truth bounding boxes, the middle column represents DEAL-YOLO with standard inference and the right column represents results of two-stage inference.}
\end{figure}

In this section, we analyze the qualitative performance of our model on the WAID and BuckTales datasets. By visualizing the predictions as shown in \ref{fig:combined}, we assess how well the model localizes animals and generalizes beyond the provided annotations. The following observations highlight key aspects of the model’s effectiveness in real-world scenarios.

The model's predictions exhibit a closer and more precise alignment with the detected animals than the ground truth annotations, demonstrating superior localization. Notably, the model also identifies animals that are missing from the ground truth labels, highlighting its ability to generalize beyond the provided annotations. The use of two-stage inference further enhances detection performance by boosting confidence scores and effectively resolving overlapping bounding boxes. This approach ensures more precise predictions and better differentiation of multiple animals within the same frame, ultimately improving overall detection accuracy.

\begin{figure}[h]
    \centering
    \includegraphics[width=5in,height=3in ]{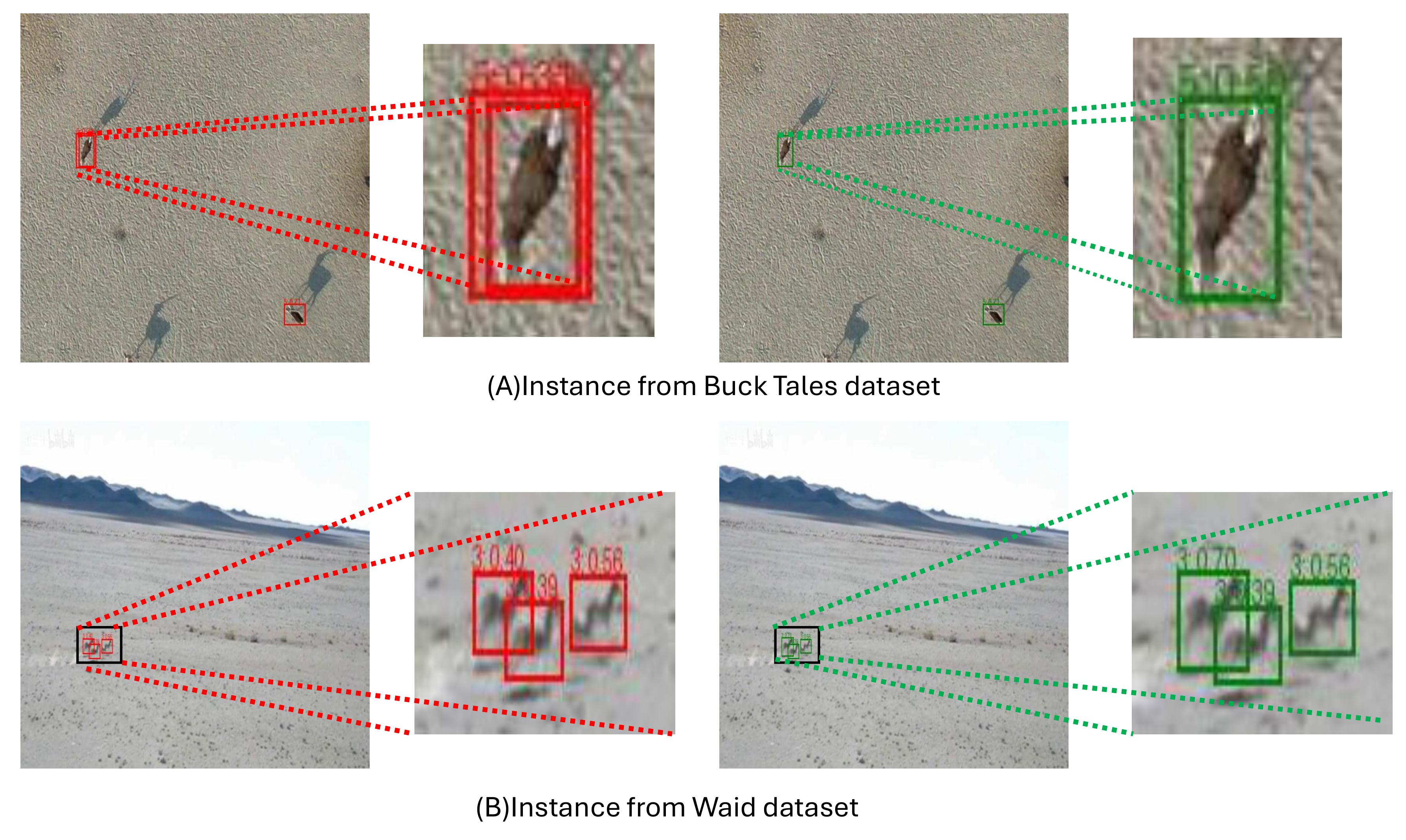} \\
    \label{fig:annot}
    \caption{Comparing the ROI of predicted anchor boxes from a single inference (shown in red) versus a two-step inference (shown in green), highlighting the removal of overlapping boxes and the increase in object confidence scores.}
    
\end{figure}

During our analysis, we identified a potential issue in both datasets. DEAL-YOLO, when using standard inference, achieves higher confidence scores than vanilla YOLOv8. As shown in Fig. \ref{fig:combined}, certain instances in both datasets lack highly accurate bounding boxes. Visualizing DEAL-YOLO's results revealed that two-stage inference produced bounding boxes that were more compact and closely fitted than the provided labels. As shown in Fig. \ref{fig:annot}, zooming into the ROI further illustrates the advantages of two-stage inference over single-stage inference. The authors recognize this as an open problem and plan to explore its implications in drone surveillance, performance quantification, and wild animal detection in future work.

\end{document}